# Analog and Multi-modal Manufacturing Datasets Acquired on the Future Factories Platform V2

Ramy Harik, Fadi El Kalach, Jad Samaha, Philip Samaha, Devon Clark, Drew Sander, Liam Burns, Ibrahim Yousif, Victor Gadow, Ahmed Mahmoud, Thorsten Wuest

Department of Mechanical Engineering, University of South Carolina

Columbia, South Carolina, USA, 29201

Corresponding Author: harik@mailbox.sc.edu

## Abstract

This paper presents two industry-grade datasets captured during an 8-hour continuous operation of the manufacturing assembly line at the Future Factories Lab, University of South Carolina, on 08/13/2024. The datasets adhere to industry standards, covering communication protocols, actuators, control mechanisms, transducers, sensors, and cameras. Data collection utilized both integrated and external sensors throughout the laboratory, including sensors embedded within the actuators and externally installed devices. Additionally, high-performance cameras captured key aspects of the operation. In a prior experiment [1], a 30-hour continuous run was conducted, during which all anomalies were documented. Maintenance procedures were subsequently implemented to reduce potential errors and operational disruptions. The two datasets include: (1) a time-series analog dataset, and (2) a multi-modal time-series dataset containing synchronized system data and images. These datasets aim to support future research in advancing manufacturing processes by providing a platform for testing novel algorithms without the need to recreate physical manufacturing environments. Moreover, the datasets are open-source and designed to facilitate the training of artificial intelligence models, streamlining research by offering comprehensive, ready-to-use resources for various applications and projects.

## I. Problem

The transition to Industry 4.0 presents significant challenges, demanding substantial changes to existing infrastructure. Establishing a reliable cyber-physical infrastructure requires more than just data collection. It involves leveraging data to understand the past, present, and future states of manufacturing processes and making informed decisions based on that data. As a result, the framework must be robust enough to facilitate seamless and efficient manufacturing systems.

A critical first step in building this infrastructure is the integration of advanced sensors that generate the necessary data to provide insights into the manufacturing environment. Capturing multiple aspects of the process is essential, but effectively utilizing this data is equally important. Whether through time-series data or images focusing on specific process elements, the goal is to extract actionable insights and train models to enhance operations. One of the primary challenges lies in developing and implementing physical testbeds to generate such data, which are crucial for designing specialized tools and advancing manufacturing processes.

The availability of open-source manufacturing data remains a significant challenge, primarily due to the proprietary nature of such data. Most companies are reluctant to share their datasets, as they are often critical to maintaining competitive advantages. This is further complicated by the inherent complexity of industrial processes, making it difficult to create comprehensive and descriptive datasets that accurately represent these operations.

Developing effective models to predict anomalies or malfunctions requires capturing sensor data during such events. However, introducing controlled anomalies in functioning industrial facilities is impractical, as disruptions can result in downtime that may take days to recover from. Consequently, the opportunity to collect open-source data from active industrial environments is limited.

Additionally, the vast scale of data generated in manufacturing processes demands high-performance computing resources for analysis. This adds to the difficulty of collecting, processing, and sharing datasets, further contributing to the scarcity of open-source manufacturing data for research and development.

To address these challenges, a modular approach to data collection offers a promising solution. The Future Factories Lab at the University of South Carolina has developed a modular manufacturing environment built to meet industrial standards. The lab is equipped with advanced sensors integrated across all actuators to capture critical information about the manufacturing processes. Additionally, multiple types of cameras are deployed to monitor and record essential aspects of operations.

The collected data supports various research topics aimed at advancing Smart Manufacturing and enabling the adoption of Industry 4.0. This paper presents an overview of the laboratory, detailing the techniques used for data communication and recording. The datasets are designed to be modular, facilitating seamless use by other researchers working on diverse topics within the industry 4.0 framework.

# II. Previous Dataset

In the previously published dataset, two types of data were recorded and presented. The first is the analog dataset, consisting of time-series datasets collected from various sensors integrated throughout the lab. The second is the multi-modal dataset, which includes a broad set of data sources synchronized with images captured from two cameras strategically positioned to extract important aspects. The data was gathered by running the manufacturing process continuously for 30 hours, during which 325 complete cycles of assembling and disassembling a rocket model were recorded.

To introduce anomalies for data collection and develop systems capable of detecting and addressing such issues, intentional defects were introduced during the experiment. These defects were categorized into three types:

1. **NoNoseCone**: Removal of the nose part
2. **NoBody2, NoNose**: Removal of the nose and the second middle part
3. **NoBody1, NoBody2, NoNose**: Removal of the nose and both middle parts

The multi-modal dataset captures a broad range of data tags from the same assembly and disassembly cycles, along with synchronized images. A total of 166,000 data samples were recorded throughout the 30-hour experiment.

The analog dataset was stored in CSV files, as it contains time-series data. The multi-modal data, however, was recorded differently. Images from the two cameras were organized into folders, with each folder containing 1,000 images per batch. Corresponding data samples were stored in JSON files, with each file representing the metadata of the related image batch. This structure resulted in 166 image batches, each accompanied by its corresponding JSON file.

# III. Current Dataset

Following the collection and analysis of the previous dataset, corrective actions were implemented to address identified issues, and another iteration of the manufacturing process was conducted to assess the effectiveness of these measures. Additionally, a new dataset was introduced, which also focuses on worker safety within the factory environment. In this iteration, the manufacturing process operated for 8 hours, during which 93 complete cycles were recorded. While this dataset shares structural similarities with the previous one, it incorporates new features and addresses distinct challenges encountered during the process.

The analog dataset remains a time-series dataset derived from various sensors throughout the lab. A key enhancement in this iteration is the inclusion of a cycle state feature, which provides insight into the specific phase of the manufacturing cycle at any given time. The full list and description of each cycle state can be found in Appendix A. However, a critical issue emerged during this run: the potentiometer on robot R02 failed at timestamp “18:19:25.029”, causing its value to drop to zero. Beyond that point, only noise signals were captured from this sensor, highlighting a sensor failure that will need to be addressed.

The multi-modal dataset retains the same structure as in the previous iteration, with 85,000 data points collected. As before, images were captured in batches of 1,000, with each batch accompanied by a JSON file that records the metadata for the corresponding images**. This iteration introduces a new focus on worker safety, where workers, equipped with various safety gear, roamed within the cell. The sequence of images, from ID “072125_1” to “078071_1”, captures these workers as they move through the environment.**

The safety equipment used in this experiment included a helmet, safety goggles, and a safety vest. To introduce variability, the workers changed their positions and poses, periodically removing one or two pieces of safety gear. The goal of this dataset is to serve as a training resource for computer vision models designed to detect whether workers are properly equipped with all required safety gear.

# IV. Experimental Setup

Although this dataset includes an additional dataset, the same infrastructure was used for data collection. Please refer to the following link for detailed information on the setup: **https://arxiv.org/pdf/2401.15544**

# V. Data Metrics

## a) Analog Dataset

As previously outlined, the dataset captures an 8-hour manufacturing run involving the assembly and disassembly of a rocket composed of four parts. To simulate potential defect scenarios, intentional anomalies were introduced by removing specific components during the process, reflecting real-world manufacturing challenges. As mentioned earlier, the dataset includes an additional feature—the cycle state—which provides insight into the current phase of the assembly and disassembly cycle.

This dataset is structured as a time-series dataset containing the following data files, as illustrated in Figure 1:

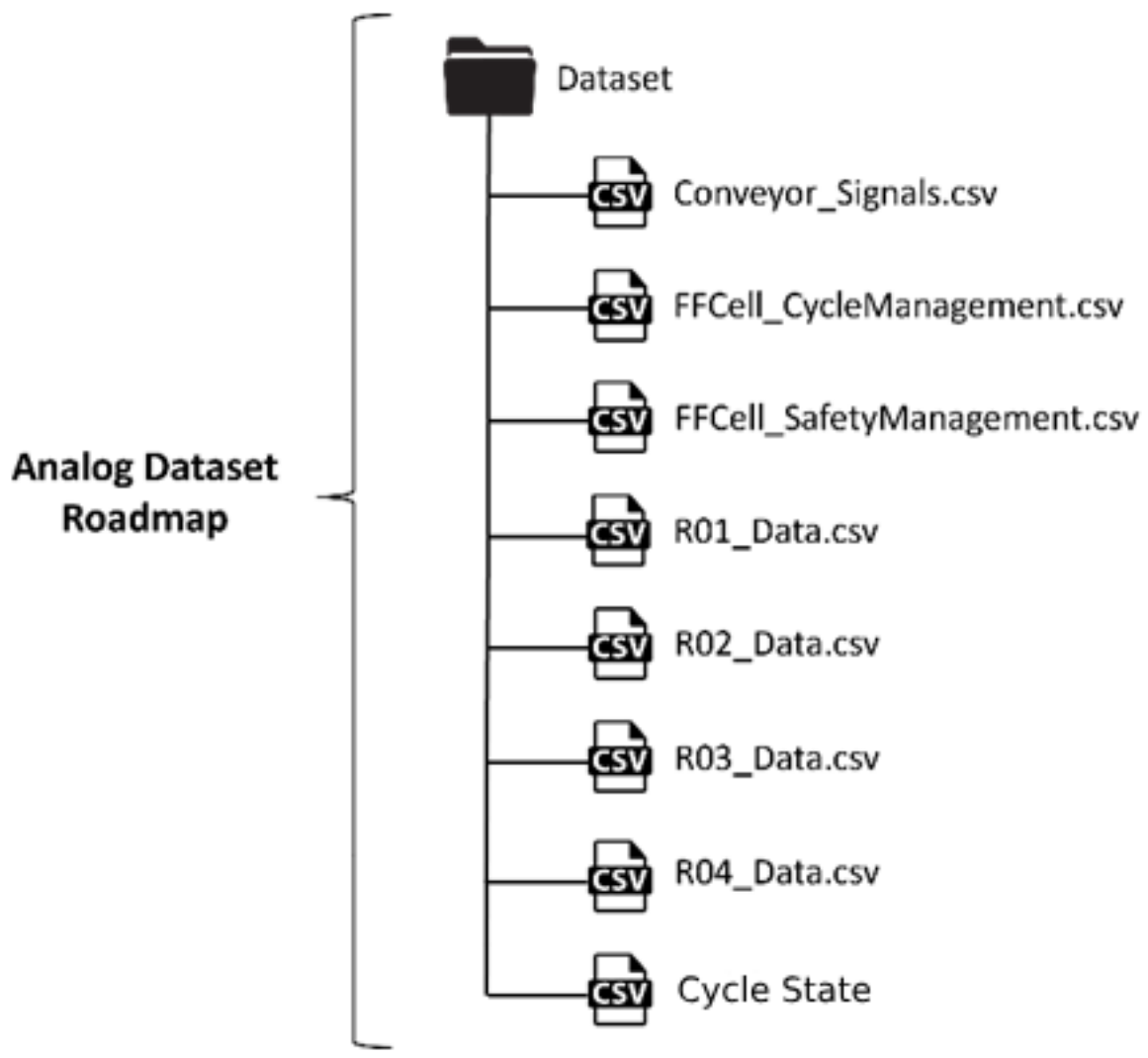

*Figure 1. Analog Data Structure*

## b) Multi-Modal Dataset

The multi-modal dataset consists of 85,000 data points collected over an 8-hour run. For each data point, two images were captured simultaneously by two separate cameras positioned within the lab. The dataset is organized into folders, with each folder containing a batch of 1,000 instances. Accompanying each folder is a JSON file that stores the synchronized data corresponding to the images within that batch.

In the multi-modal dataset, the load cell readings were converted to pound-force, enhancing the ease of data interpretation for the user. The conversion was performed using the following formula:

$$Load_{lbf} = \frac{(Load\ Cell\ Value) - 1000}{14000} * 25$$

As previously mentioned, this iteration introduces an additional feature focusing on the safety realm, capturing scenarios related to worker safety. The structure of the multi-modal dataset is illustrated in Figure 2:

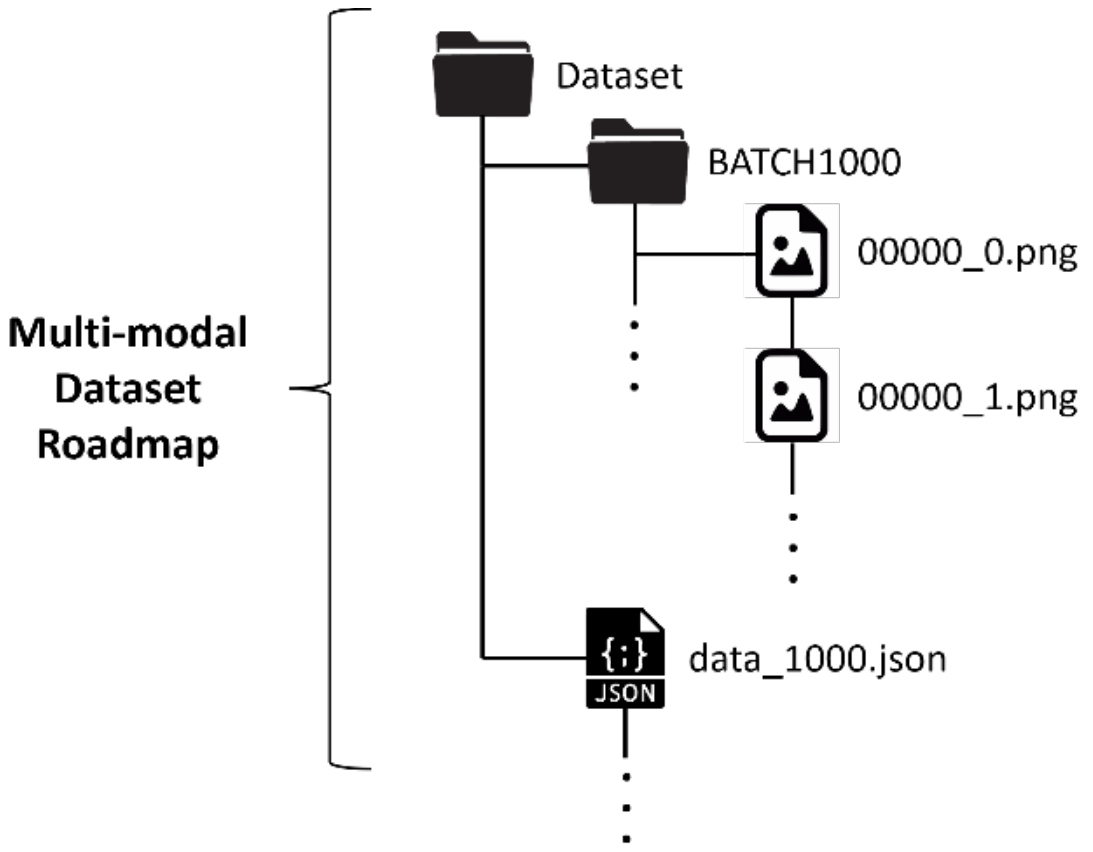

*Figure 2. Multi-Modal Data Structure*

## c) Anomalies

The previously discussed anomalies were systematically classified and enumerated. The identified anomaly classes include:

- **"No Nose"**
- **"No Nose and No Body 2"**
- **"No Nose, No Body 2, and No Body 1"**
- **"Normal"**

The distribution of these anomaly classes is visualized in figure 3 below, providing an overview of the frequency and occurrence of each class within the dataset.

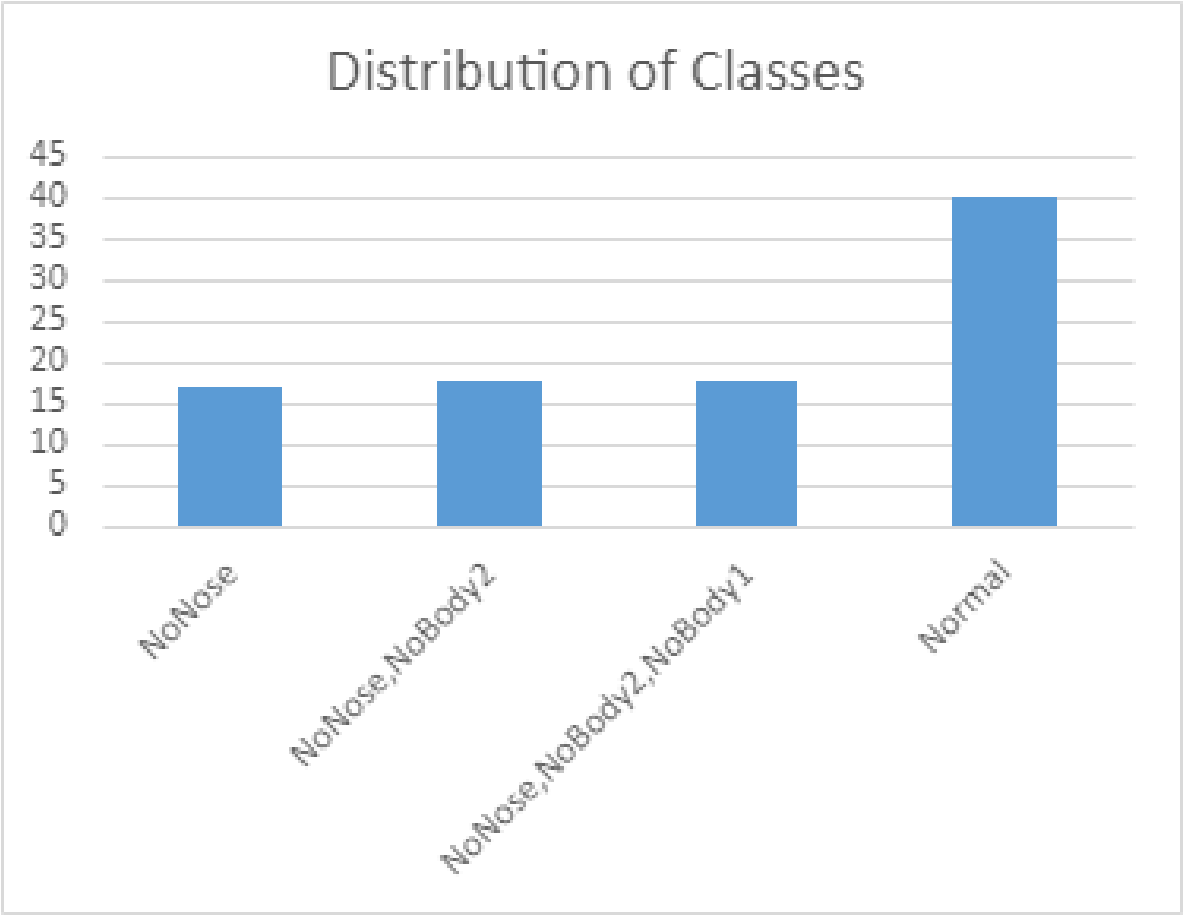

*Figure 3. Anomaly Classes Distribution*

## Download Links:

The **analog dataset** is downloadable using the following link:

- Analog Link 1 [https://www.kaggle.com/datasets/ramyharik/ff-2024-08-13-analog-dataset]

The **multi-modal dataset** is divided into 3 different downloadable links:

- Multi-Modal Link 1 [https://www.kaggle.com/datasets/ramyharik/ff-2024-08-13-multi-modal-dataset-13]
- Multi-Modal Link 2 [https://www.kaggle.com/datasets/ramyharik/ff-2024-08-13-multi-modal-dataset-23]
- Multi-Modal Link 3 [https://www.kaggle.com/datasets/ramyharik/ff-2024-08-13-multi-modal-dataset-33]

## Acknowledgements

This work is funded in part by NSF Award 2119654 "RII Track 2 FEC: Enabling Factory to Factory (F2F) Networking for Future Manufacturing," and "Enabling Factory to Factory (F2F) Networking for Future Manufacturing across South Carolina," funded by South Carolina Research Authority. Any opinions, findings, conclusions, or recommendations expressed in this material are those of the author(s) and do not necessarily reflect the views of the sponsors.

## License

"Creative Commons Attribution-ShareAlike 4.0 International Public License" is included with the data. Under this license,

- Users can share the dataset or any publication that uses the data by giving credit to the data provider. The user must cite this paper for the credit.
- Users can distribute any additions, transformations, or changes to your dataset under this license. However, the same license needs to be added to any redistributed data. Hence, any user of the adapted dataset would likewise need to share their work with this license.

## Appendix A

| Cycle State | Description of Cycle State |
|---|---|
| 1 | R01 Picks Tray from MHS |
| 2 | R01 Places Tray on Conveyor |
| 3 | R01 Back to Home Position and Conveyors On |
| 4 | R02 Pick Body 1 from Conveyor |
| 5 | R02 Place Body 1 on local station |
| 6 | R02 Pick Body 2 from Conveyor |
| 7 | R02 Place Body 2 on local station |
| 8 | R02 and R03 Assemble Rocket Together |
| 9 | Conveyors move assembled rocket to R04 and R04 picks up tray |
| 10 | R04 place tray on fixture |
| 11 | R04 Disassemble Nose |
| 12 | R04 Disassemble Body 2 |
| 13 | R04 Disassemble Body 1 |
| 14 | R04 Disassemble Tail |
| 15 | R04 Place Tail Back on Tray |
| 16 | R04 Place Nose Back on Tray |
| 17 | R04 Place Body 1 Back on Tray |
| 18 | R04 Place Body 2 Back on Tray |
| 19 | R04 Pick Disassembled Tray |
| 20 | R04 Place Tray on MHS |
| 21 | R04 Back to Home Position |

# Appendix B

| Asset | Sensor Values | Data Type | Multi-Modal Dataset | Analog Dataset | Description |
|---|---|---|---|---|---|
| Conveyors | Q_VFD1_Temperature | Float | ✓ | ✓ | The temperature of conveyor 1 in Fahrenheit |
| | Q_VFD2_Temperature | Float | ✓ | ✓ | The temperature of conveyor 2 in Fahrenheit |
| | Q_VFD3_Temperature | Float | ✓ | ✓ | The temperature of conveyor 3 in Fahrenheit |
| | Q_VFD4_Temperature | Float | ✓ | ✓ | The temperature of conveyor 4 in Fahrenheit |
| | M_Conv1_Speed_mmps | Integer | ✓ | | The speed of the conveyor 1 in mm/s |
| | M_Conv2_Speed_mmps | Integer | ✓ | | The speed of the conveyor 2 in mm/s |
| | M_Conv3_Speed_mmps | Integer | ✓ | | The speed of the conveyor 3 in mm/s |
| | M_Conv4_Speed_mmps | Integer | ✓ | | The speed of the conveyor 4 in mm/s |
| Grippers | I_R01_Gripper_Pot | Integer | ✓ | ✓ | The analog output signal of the potentiometer on the Robot 1 gripper |
| | I_R02_Gripper_Pot | Integer | ✓ | ✓ | The analog output signal of the potentiometer on the Robot 2 gripper |
| | I_R03_Gripper_Pot | Integer | ✓ | ✓ | The analog output signal of the potentiometer on the Robot 3 gripper |
| | I_R04_Gripper_Pot | Integer | ✓ | ✓ | The analog output signal of the potentiometer on the Robot 4 gripper |
| | I_R01_Gripper_Load | Integer | ✓ | ✓ | The analog output signal of the load cell on the Robot 1 gripper |
| | I_R02_Gripper_Load | Integer | ✓ | ✓ | The analog output signal of the load cell on the Robot 2 gripper |
| | I_R03_Gripper_Load | Integer | ✓ | ✓ | The analog output signal of the load cell on the Robot 3 gripper |
| | I_R04_Gripper_Load | Integer | ✓ | ✓ | The analog output signal of the load cell on the Robot 4 gripper |
| | I_R01_Gripper_Load_lbf | Integer | ✓ | | The analog output signal of the load cell on the Robot 1 gripper converted to lbf |
| | I_R02_Gripper_Load_lbf | Integer | ✓ | | The analog output signal of the load cell on the Robot 2 gripper converted to lbf |
| | I_R03_Gripper_Load_lbf | Integer | ✓ | | The analog output signal of the load cell on the Robot 3 gripper converted to lbf |
| | I_R04_Gripper_Load_lbf | Integer | ✓ | | The analog output signal of the load cell on the Robot 4 gripper converted to lbf |
| Robot 1 | M_R01_SJointAngle_Degree | Float | ✓ | ✓ | The joint S angle of Robot 1 in degrees |
| | M_R01_LJointAngle_Degree | Float | ✓ | ✓ | The joint L angle of Robot 1 in degrees |
| | M_R01_UJointAngle_Degree | Float | ✓ | ✓ | The joint U angle of Robot 1 in degrees |
| | M_R01_RJointAngle_Degree | Float | ✓ | ✓ | The joint R angle of Robot 1 in degrees |
| | M_R01_BJointAngle_Degree | Float | ✓ | ✓ | The joint B angle of Robot 1 in degrees |
| | M_R01_TJointAngle_Degree | Float | ✓ | ✓ | The joint T angle of Robot 1 in degrees |
| Robot 2 | M_R02_SJointAngle_Degree | Float | ✓ | ✓ | The joint S angle of Robot 2 in degrees |
| | M_R02_LJointAngle_Degree | Float | ✓ | ✓ | The joint L angle of Robot 2 in degrees |
| | M_R02_UJointAngle_Degree | Float | ✓ | ✓ | The joint U angle of Robot 2 in degrees |
| | M_R02_RJointAngle_Degree | Float | ✓ | ✓ | The joint R angle of Robot 2 in degrees |
| | M_R02_BJointAngle_Degree | Float | ✓ | ✓ | The joint B angle of Robot 2 in degrees |
| | M_R02_TJointAngle_Degree | Float | ✓ | ✓ | The joint T angle of Robot 2 in degrees |
| Robot 3 | M_R03_SJointAngle_Degree | Float | ✓ | ✓ | The joint S angle of Robot 3 in degrees |
| | M_R03_LJointAngle_Degree | Float | ✓ | ✓ | The joint L angle of Robot 3 in degrees |
| | M_R03_UJointAngle_Degree | Float | ✓ | ✓ | The joint U angle of Robot 3 in degrees |
| | M_R03_RJointAngle_Degree | Float | ✓ | ✓ | The joint R angle of Robot 3 in degrees |
| | M_R03_BJointAngle_Degree | Float | ✓ | ✓ | The joint B angle of Robot 3 in degrees |
| | M_R03_TJointAngle_Degree | Float | ✓ | ✓ | The joint T angle of Robot 3 in degrees |
| Robot 4 | M_R04_SJointAngle_Degree | Float | ✓ | ✓ | The joint S angle of Robot 4 in degrees |
| | M_R04_LJointAngle_Degree | Float | ✓ | ✓ | The joint L angle of Robot 4 in degrees |
| | M_R04_UJointAngle_Degree | Float | ✓ | ✓ | The joint U angle of Robot 4 in degrees |
| | M_R04_RJointAngle_Degree | Float | ✓ | ✓ | The joint R angle of Robot 4 in degrees |
| | M_R04_BJointAngle_Degree | Float | ✓ | ✓ | The joint B angle of Robot 4 in degrees |
| | M_R04_TJointAngle_Degree | Float | ✓ | ✓ | The joint T angle of Robot 4 in degrees |
| Safety | I_SafetyDoor1_Status | Bool | ✓ | ✓ | "True" if Safety Door 1 is open and "False" if otherwise |
| | I_SafetyDoor2_Status | Bool | ✓ | ✓ | "True" if Safety Door 2 is open and "False" if otherwise |
| | I_HMI_EStop_Status | Bool | | ✓ | "True" if the HMI E-Stop button has been pressed and "False" if otherwise |
| Cycle Management | Q_Cell_CycleCount | Integer | ✓ | ✓ | Integer value representing the number of cycles elapsed. **Note:** This number resets to zero whenever the cycle was interrupted |
| | Q_Cell_CycleState | Integer | ✓ | ✓ | Integer value representing the specific phase in the assembly cycle the system is currently in |
| Material Handling Station | I_MHS_GreenRocketTray | Bool | ✓ | ✓ | "True" if the Green Rocket Tray is detected in the Material Handling Station and "False" if otherwise |
| Stopper | I_Stopper1_Status | Bool | | ✓ | "True" if Stopper 1 is extended and "False" if otherwise |
| | I_Stopper2_Status | Bool | | ✓ | "True" if Stopper 2 is extended and "False" if otherwise |
| | I_Stopper3_Status | Bool | | ✓ | "True" if Stopper 3 is extended and "False" if otherwise |
| | I_Stopper4_Status | Bool | | ✓ | "True" if Stopper 4 is extended and "False" if otherwise |
| | I_Stopper5_Status | Bool | | ✓ | "True" if Stopper 5 is extended and "False" if otherwise |
| Cameras | Path1 | String | ✓ | | Path to image taken from Camera 1 |
| | Path2 | String | ✓ | | Path to image taken from Camera 2 |